\documentclass[letterpaper]{article} % DO NOT CHANGE THIS
\usepackage{aaai25}  % DO NOT CHANGE THIS
\usepackage{times}  % DO NOT CHANGE THIS
\usepackage{helvet}  % DO NOT CHANGE THIS
\usepackage{courier}  % DO NOT CHANGE THIS
\usepackage[hyphens]{url}  % DO NOT CHANGE THIS
\usepackage{graphicx} % DO NOT CHANGE THIS
\usepackage{natbib}  % DO NOT CHANGE THIS AND DO NOT ADD ANY OPTIONS TO IT
\usepackage{caption} % DO NOT CHANGE THIS AND DO NOT ADD ANY OPTIONS TO IT
\usepackage{algorithm}
\usepackage{algorithmic}

\usepackage{newfloat}
\usepackage{listings}
\DeclareCaptionStyle{ruled}{labelfont=normalfont,labelsep=colon,strut=off} % DO NOT CHANGE THIS
\floatstyle{ruled}
\newfloat{listing}{tb}{lst}{}
\floatname{listing}{Listing}
\def\ie{\emph{i.e}\onedot}

\def\etal{et al\onedot}
\usepackage{xspace}
\usepackage{graphicx}
\usepackage{amsmath}
\usepackage{amssymb}
\usepackage{soul}
\usepackage{booktabs}
\usepackage{comment}
\usepackage{multirow}
\usepackage{bm}
\usepackage{subcaption}
\newcommand{\new}[1]{{#1}}

\newcommand{\aaai}[1]{{#1}}
\usepackage{tabularx}

\newcommand{\tool}{{QORT-Former}\xspace}
\title{\tool: Query-optimized Real-time Transformer for Understanding Two Hands Manipulating Objects}

\author{Elkhan Ismayilzada\textsuperscript{\rm 1, \rm 2}\equalcontrib\thanks{This work was conducted when Elkhan Ismayilzada was graduate student at UNIST.}, MD Khalequzzaman Chowdhury Sayem\textsuperscript{\rm 1}\equalcontrib, Yihalem Yimolal Tiruneh\textsuperscript{\rm 1}, Mubarrat Tajoar Chowdhury\textsuperscript{\rm 1}, Muhammadjon Boboev\textsuperscript{\rm 1}, Seungryul Baek\textsuperscript{\rm 1}\thanks{Corresponding author.}}

\affiliations{
    \textsuperscript{\rm 1}UNIST, Ulsan, South Korea\\
    \textsuperscript{\rm 2}Michigan State University, MI, USA\\
    \{elkhan, khalequzzamansayem, yihalemyimolal, mubarrattajoar, muhammad, srbaek\}@unist.ac.kr
    
}

\begin{document}

\maketitle
\begin{abstract}
Significant advancements have been achieved in the realm of 
understanding poses and interactions of two hands manipulating an object. The emergence of augmented reality (AR) and virtual reality (VR) technologies has heightened the demand for real-time performance in these applications. However, current state-of-the-art models often exhibit promising results at the expense of substantial computational overhead. In this paper, we present a query-optimized real-time Transformer (\tool), the first Transformer-based real-time framework for 3D pose estimation of two hands and an object. We first limit the number of queries and decoders to meet the efficiency requirement. Given limited number of queries and decoders, \new{we propose to optimize queries which are taken as input to the Transformer decoder, to secure better accuracy:} (1) we propose to divide queries into three types (a left hand query, a right hand query and an object query) and enhance query features (2) by using the contact information between hands and an object and (3) by using three-step update of enhanced image and query features with respect to one another.
With proposed methods, we achieved real-time pose estimation performance using just 108 queries and 1 decoder (53.5 FPS on an RTX 3090TI GPU). Surpassing state-of-the-art results on the H2O dataset by 17.6\% (left hand), 22.8\% (right hand), and 27.2\% (object), as well as on the FPHA dataset by 5.3\% (right hand) and 10.4\% (object), our method excels in accuracy. Additionally, it sets the state-of-the-art in interaction recognition, maintaining real-time efficiency with an off-the-shelf action recognition module.
\begin{links}
    \link{Project Page}{https://kcsayem.github.io/QORT-Former/}
\end{links}
\end{abstract}
\section{Introduction}
\label{sec:intro} 
\begin{figure}[t!]
\centering
\includegraphics[width=\linewidth]{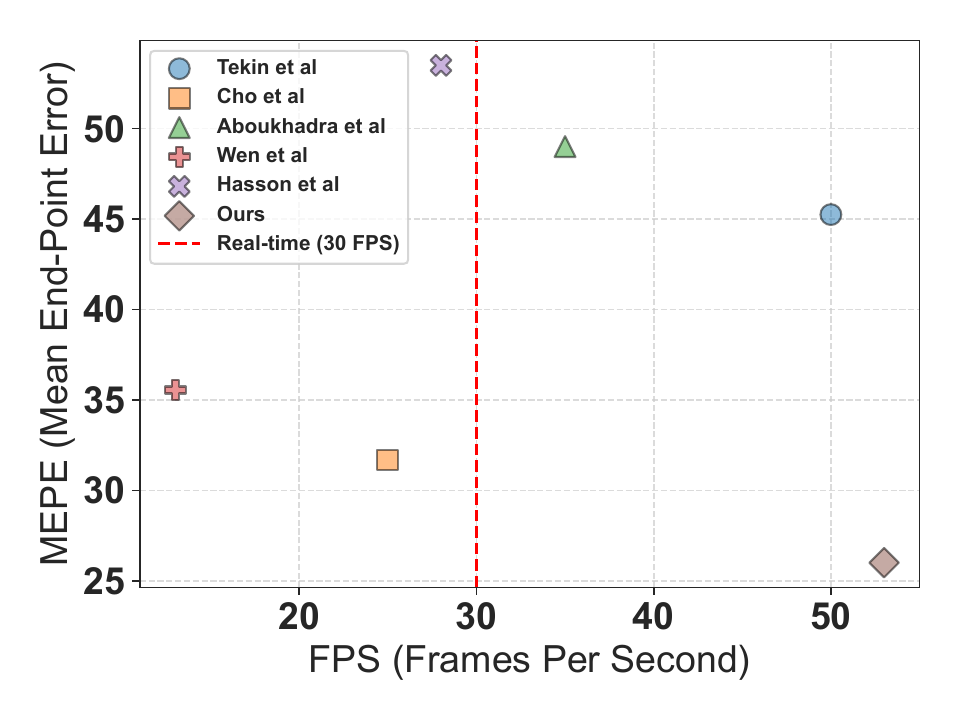}
\caption{Comparisons to competitive state-of-the-art algorithms~\cite{cho2023transformer, tekin2019h+, aboukhadra2022thor, wang2023interacting, hasson2020leveraging} on the two hands and an object pose estimation task on an RTX 3090TI GPU. Even with the Transformer architecture, we achieved the fastest speed (53.5 FPS) while obtaining the best accuracy among the methods.}
\label{fig:ablation}
\end{figure}
Estimating poses and actions in egocentric videos involving two hands and an object is crucial for applications like AR, VR, and HCI. Significant progress has been made in hand pose estimation \cite{zimmerman_iccv2017, baek2018augmented, baek2019pushing, baek2020weakly, kim2021end, garcia2018first, wang2020rgb2hands, moon2020interhand2, lin2021mesh, chen2021temporal, chen2022mobrecon, lee2023image} and object 6D pose estimation \cite{xiang2017posecnn, li2019cdpn, tekin2018real, kehl2017ssd, chen2020learning, iwase2021repose, xu2022rnnpose}, which were previously addressed independently. Given that hands frequently interact with objects, there's a growing demand for methods that jointly estimate the poses of both hands and objects, along with hand interaction classes \cite{hasson2020leveraging, hasson2019learning, armagan2020measuring, liu2021semi, cho2023transformer, fan2024hold, fan2025benchmarks}. Recent frameworks like H2OTR \cite{cho2023transformer} have shown impressive performance in estimating two-hand and object poses, object types, and hand interaction classes in a single Transformer-based framework. However, H2OTR's use of the deformable DETR architecture \cite{zhu2020deformable} for frame-by-frame pose estimation leads to significant computational overhead, making it unsuitable for real-time applications, since pose estimation accounts for over 97\% of the total inference time.\\
This paper aims to achieve real-time 3D pose estimation for two hands and an object by improving both computational efficiency and accuracy. In deformable DETR \cite{zhu2020deformable}, the encoder consumes 49\% of the overall GFLOPs while contributing only 11\% of the AP \cite{lin2022d}, mainly due to the multi-scale deformable attention mechanism and numerous decoder layers. Inspired by recent efficient encoder designs \cite{cheng2022sparse, he2023fastinst, lv2023detrs}, we developed a feature decoder based on the pyramid pooling module (PPM) \cite{zhao2017pyramid} to expand receptive fields and reduce computational costs. We also optimized the number of queries by introducing hand-object queries tailored for estimating poses, explicitly utilizing 2D locations of hands and objects. To minimize reliance on heavy feature decoder, we update enhanced features and query features within the Transformer decoder. Unlike H2OTR \cite{cho2023transformer}, which randomly initializes object queries, we select queries with high semantics from multi-scale feature maps. Despite using fewer queries (108 compared to H2OTR's 300), our method yields higher-quality queries, enhancing both speed and accuracy (Refer to Fig \ref{fig:queries}).\\
An essential facet of two hands and an object pose estimation is the complex phenomenon of grasping, which involves intricate hand configurations and contact regions between the hands and the object. Despite progress in estimating hand-object interaction poses, identifying contact points remains challenging. To address this, we estimate a contact map from the feature decoder’s output and integrate it into our query features before passing them to the Transformer decoders. \\
 To reduce dependence on a heavy feature decoder, in our proposed decoder both image and query features are co-optimized for enhanced hand-object pose estimation. We combine query features for the hands and object with auxiliary background queries. Unlike traditional methods that update only query features, our decoder refines image features and query features through a three-step process: 1) cross-attention improves spatial and contextual relationships, 2) location-based enhancement focuses on key areas around the hands and object, and 3) further refinement captures fine details like finger joints and contact points, leading to more accurate pose and class estimation.\\
Our proposed modification ensures superior performance with just $100$ hand-object queries (~$+$~$8$ auxiliary queries) and $1$ decoder, compared to the $300$ randomly initialized queries and $6$ decoders in H2OTR~\cite{cho2023transformer}; which enables us to achieve real-time performance at 53.5 FPS, significantly outperforming the 26 FPS in H2OTR~\cite{cho2023transformer} on an RTX 3090TI GPU. We also demonstrate the state-of-the-art two hands and an object pose and action recognition performance on H2O \cite{kwon2021h2o} and FPHA \cite{garcia2018first} datasets. To summarize, our main contributions are as follows:
\begin{itemize}
    \item We present the query-optimized real-time Transformer~(\tool), to the best of our knowledge, the first Transformer-based real-time framework for 3D pose estimation of two hands and an object.

    \item For the real-time speed, we proposed to constrain the query numbers (as $108$) and the number of decoders (as $1$). See Figure \ref{fig:ablation} for FPS vs. Error comparison with other methods.
    
    \item For robust accuracy with a reduced number of queries and decoders, we propose a novel method of dividing object queries into three sections: a left hand, a right hand, and an object to optimize the location of queries. We also introduce the incorporation of contact map features into query features, enhancing the query's awareness of contact dynamics in two hands and an object interactions.

    \item To reduce the dependency on heavy feature decoders, we introduce a three-step feature update in the transformer decoder, simultaneously constraining the decoder count.
    
    \item Our proposed method \new{outperforms current state-of-the-art by an impressive margin (5.3\%-27.2\%)} in pose estimation on H2O~\cite{kwon2021h2o} and FPHA~\cite{garcia2018first} datasets while ensuring real-time performance (53.5 FPS on an RTX 3090TI GPU).
\end{itemize}
\section{Related Works}
\label{sec:relatedworks}
In this section, we discuss the previous related works in the domain of hand-object pose estimation. Building upon the success of transformers \cite{vaswani2017attention} and the subsequent emergence of ViT \cite{dosovitskiy2020image}, numerous transformer-based methodologies \cite{carion2020end, wang2022anchor, yao2021efficient, li2022dn, liu2022dab, cha2024text2hoi} have been successfully applied across multiple vision-related tasks \cite{han2022survey}, including hand pose estimation \cite{huang2020hand, Jiang_2023_a2j-tran, Fu_2023_ICCV, Zhang_2024_WACV, pavlakos2024reconstructing} and hand-object pose estimation \cite{hampali2022keypoint,liu2021semi, cho2023transformer}.
\citeauthor{hampali2022keypoint} introduced a transformer-based 3D hand-object pose estimation methodology that performs self-attention between 2D hand-keypoint features. \citeauthor{Fu_2023_ICCV} propose deformer, a dynamic fusion Transformer that leverages spatial relationships within an image and temporal correlations between nearby frames to learn hand deformations. More recently, A2J-Transformer \cite{Jiang_2023_a2j-tran} extends the state-of-the-art depth-based 3D single hand pose estimation method A2J \cite{Xiong_2019_a2j} to the RGB domain under interacting hand conditions. \cite{Jiang_2023_a2j-tran} enhances A2J \cite{Xiong_2019_a2j} by incorporating non-local encoding-decoding framework of transformers, enabling global spatial context awareness and adaptive feature learning for each anchor point located in 3D space.
\citeauthor{cho2023transformer} introduce a Transformer-based unified framework to estimate the poses of two hands and an object, and their interaction classes in a single inference step. Although this model is state-of-the-art in terms of accuracy, it does not perform in real-time. The major drawback in terms of speed in \citeauthor{cho2023transformer}'s work is in the pose estimator network, which takes more than 97\% of the total inference time. This is contributed by factors such as a high number of queries, heavy encoder and the use of a large number of decoder layers. To reduce the complexity of encoder, we employ PPM-FPN \cite{cheng2022sparse} and use one feature map instead of using all three feature maps as in \cite{cho2023transformer}. But this makes our encoder less feature enriched compared to \cite{cho2023transformer}. To tackle this, we propose two simple but effective modifications. Firstly, we construct semantically meaningful queries. Which are then divided into left-hand, right-hand, and object categories, similar to \cite{hampali2022keypoint}. However, we deviate by using a dedicated query proposal network to suggest locations based on semantic relevance, eliminating the need for non-maximal suppression for reduced inference time and improved efficiency. Another significant aspect of hand-object interaction is the point of contact between hands and objects \cite{karunratanakul2020grasping, yang2021cpf}. To enrich the query, in our work, we combine the contact map features along with the semantic features to be able to catch intricate details while hands and objects are in contact. Additionally, since we utilize a single feature map, thereby creating discrepancies with conventional decoders employed in complex models \cite{cho2023transformer, Cheng_2022_CVPR}, modifications to the decoder become necessary. Therefore, we opt for a three-step image and query features co-optimization strategy in the decoder, involving cross-attention twice. While this three-step update increases the decoder's complexity, it enables achieving comparable performance with a smaller number of decoder layers compared to other models. Combining all the modifications allows us to get state-of-the-art performance and real-time inference speed. 
\begin{figure*}[t!]
\centering
\includegraphics[width=\textwidth]{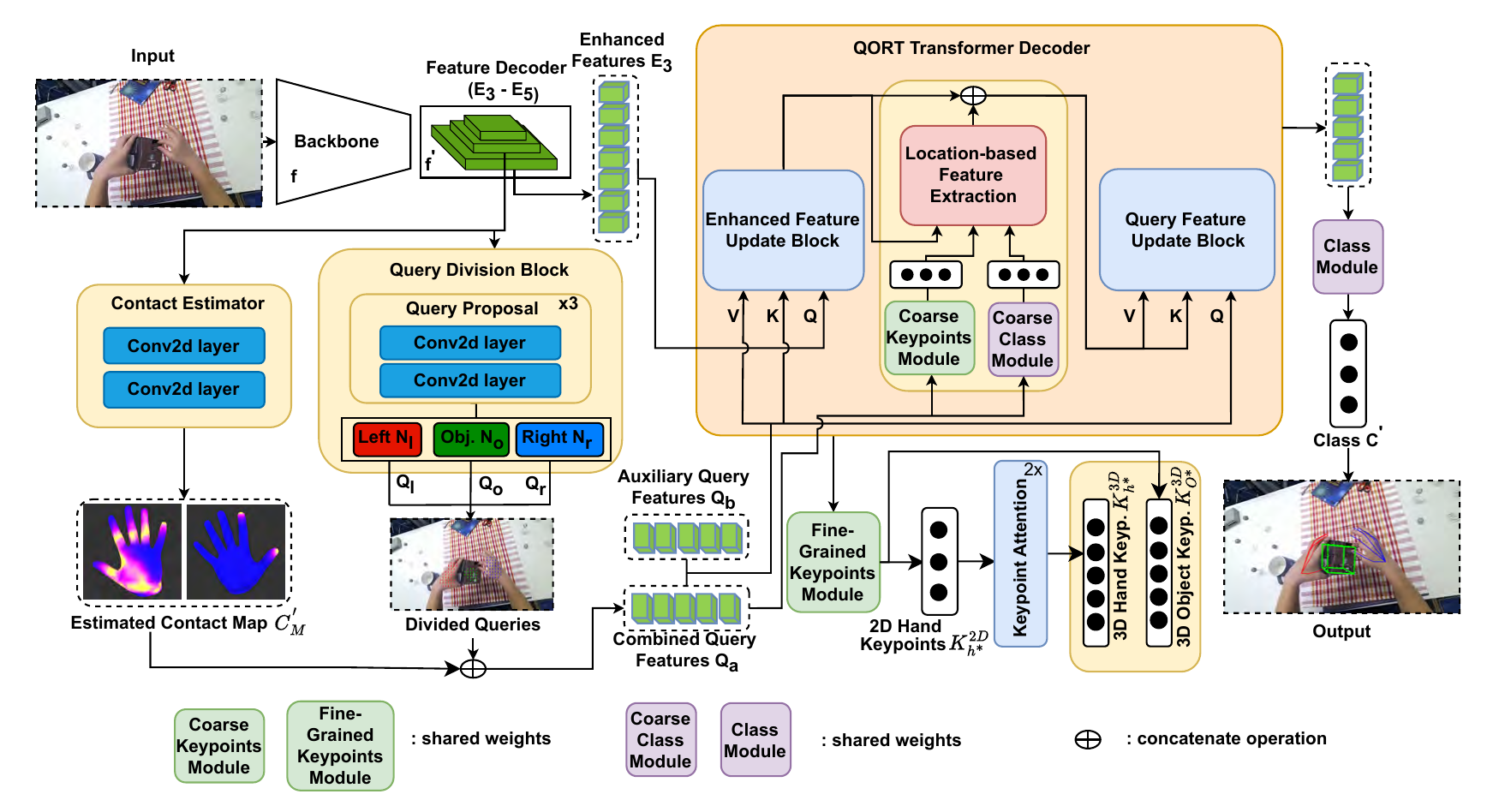}
% \resizebox{\textwidth}{!}{\includegraphics{figures/updated-arch-compressed.pdf}}
\caption{Our architecture begins with extracting a multi-scale feature $\mathbf{f}$ from an image using ResNet-50~\cite{he2016deep}, which is then refined into $\mathbf{f}'$ by our feature decoder. We propose queries aligned with hand and object locations, incorporating contact map features, while auxiliary queries capture background details. In the QORT Transformer decoder, enhanced and query features undergo three steps: 1) Cross-attention updates the enhanced feature based on integrated query features in Enhanced Feature Update Block, 2) Location-based Feature Extraction module adds feature maps of $3\times3$ patches around coarse 2D hand and object keypoints to Enhanced Feature, and 3) Cross and self-attention layers update the integrated query features based on updated enhanced features in Query Feature Update Block. Finally, the heads estimate poses for both hands and the object.}
\label{fig:Framework}
\end{figure*}
\section{Method}
We propose the hand-object interaction recognition framework that inputs an RGB image and outputs 3D poses of two hands and an object.

\subsection{Query-Optimized Real-Time Transformer}
\noindent To tackle the challenging task of recognizing 3D poses of two hands and an object, we proposed the real-time Transformer-based framework, query-optimized real-time Transformer (\tool). In this section, we will explain each component of our method in detail.

\noindent \textbf{Backbone}. Our model uses the ImageNet pre-trained ResNet-50 \cite{he2016deep} architecture to extract features from an input image $\mathbf{x}$. Specifically, we acquire three distinct feature maps $\mathbf{f}$, each of resolutions  $1/8$, $1/16$, and $1/32$ of the input image, respectively. The projection of these feature maps involves a 1$\times$1 convolutional layer, resulting in them being represented with 256 channels. Subsequently, these projected feature maps are used as input to the feature decoder for further processing. \\
\noindent \textbf{Feature Decoder}. The feature decoder takes in the projected feature maps $\mathbf{f}$ and generates enhanced multi-scale contextual feature maps $\mathbf{f}'=\{\textbf{E}_\text{3} \in \mathbb{R}^{\frac{H}{8} \times \frac{W}{8} \times 256},~\textbf{E}_\text{4} \in \mathbb{R}^{\frac{H}{16} \times \frac{W}{16} \times 256},~\textbf{E}_\text{5} \in \mathbb{R}^{\frac{H}{32} \times \frac{W}{32} \times 256}\}$ where $H$ and $W$ are the height and the width of the input image, respectively. As the feature decoder, we use the PPM-FPN \cite{cheng2022sparse}, which employs pyramid pooling module (PPM) \cite{zhao2017pyramid} to enlarge the receptive fields.\\
    \noindent \textbf{Query Division Block}. The role of object queries in Transformer architecture is paramount. Unlike H2OTR \cite{cho2023transformer}, where the object queries are randomly initialized, we propose to select queries with high semantics from underlying multi-scale feature maps by adding a \new{query proposal module} on top of the feature map to output the class probability prediction for each pixel. The output of \new{the query proposal module} is a $(N_c+1)-$dimensional probability simplex, where $N_c$ is the number of classes and one dimension is added for the ``no object'' class. We use three \new{query proposal modules}, two for classifying each hand and one for objects. As a result, each pixel is classified with three classifiers. In the context of hand-object pose estimation, classifying left and right hands is especially difficult due to the similarities in their underlying feature maps, and this \new{query proposal module} aids in reducing the uncertainty. Given the classification probability from the respective \new{module}, we adopt a strategy wherein the top $N_l$ location of the pixel corresponds to the left hand, the top $N_r$ location to the right hand, and the top $N_o$ location to objects. Here, $N_l$, $N_r$, and $N_o$ represent the number of queries designated for the left and right hands and objects, respectively. Subsequently, we generate $\bm{Q}_\text{l} \in \mathbb{R}^{N_{l} \times 256}$, $\bm{Q}_\text{r} \in \mathbb{R}^{N_{r} \times 256}$ and $\bm{Q}_\text{o} \in \mathbb{R}^{N_{o} \times 256}$ query features for each selected location utilizing our $\textbf{E}_\text{4}$ feature map from the feature decoder, where $\bm{Q}_\text{l}$, $\bm{Q}_\text{r}$, $\bm{Q}_\text{o}$ denote query features for left hand, right hand and objects. During the training phase, we implement a matching-based Hungarian~\cite{kuhn1955hungarian} loss to supervise each \new{module}. This involves leveraging class predictions along with a location cost, indicating whether the pixel is located in the region of interest for the target object. Further details regarding the loss calculation are elaborated in the Loss function section.\\
\noindent \textbf{Contact Estimator.} In the context of 3D hand-object pose estimation, a crucial aspect revolves around the nuanced dynamics of grasping—a foundational element in the interaction between two hands and an object. This complex process involves sophisticated hand configurations leading to contact zones between two hands and an object. Consequently, the integration of contact zone information into the pose estimation of interacting two hands and an object holds substantial promise for improving accuracy. With the goal of achieving the objective, our initial step involves constructing a contact map $\mathbf{C_M}\in\mathbb{R}^{2 \times 778 \times 1}$ for two hands encoding the vertex regions close to $1$ if they are contacting with an object, following the approach outlined in \cite{cho2023transformer}. For contact map estimation, we utilize the mid-sized (\ie, $\textbf{E}_\text{4}$ in Fig.~\ref{fig:Framework}.) feature map from the feature decoder. Opting for the mid-sized feature map aims to balance computational efficiency with information retention. Further details regarding the loss calculation for the contact estimator can be found in Loss function section. Once the contact maps of two hands are estimated, we add them as the contact map feature to object queries, to further improve the semantics of integrated query features.\\
\noindent \textbf{QORT Transformer Decoder}. Upon obtaining query features $\bm{Q}_\text{l}$, $\bm{Q}_\text{r}$, and $\bm{Q}_\text{o}$ with the contact map features, we concatenate them to form combined query features, $\bm{Q}_\text{a} \in \mathbb{R}^{(N_{l}+N_r+N_o) \times 256}$. These are then integrated with $\bm{Q}_\text{b}  \in \mathbb{R}^{N_b \times 256}$ auxiliary query features, where $N_b$ is the number of auxiliary queries, strategically designed to facilitate the aggregation of background features and provide general image-independent cues during the update process. The integrated queries are fed into the QORT Transformer
decoder alongside the flattened enhanced features, $\textbf{E}_3$. \\
Unlike traditional Transformer architectures \cite{cho2023transformer, Cheng_2022_CVPR}, where only query features are updated, our proposed decoder co-optimizes both image features (Enhanced Features, $\textbf{E}_3$) and query features. This reduces reliance on heavy encoders and allows for a lightweight feature decoder. The co-optimization process involves three steps:
\begin{itemize}
    \item Enhanced Feature ($\textbf{E}_3$) Update: Enhanced features and integrated query features undergo cross-attention in the ``Enhanced Feature Update Block'', where enhanced features act as queries and integrated query features as keys and values. This refines the holistic representation of the spatial configuration and contextual relationships between the hands and the object in the scene. 
    \item Location-Based Enhancement:  The enhanced feature is further refined to focus more on the areas around the hands and object. First, coarse 2D keypoints and probability simplex for each class are estimated using the ``coarse keypoints module" and ``coarse class module" from the combined query features, $Q_a$. The keypoints with the highest probability for both hands and the object are then fed into the "Location-based feature extraction" module along with the updated $\textbf{E}_3$. This module generates feature maps of $3 \times 3$ patches around the coarse 2D keypoints, which are then concatenated with the updated $\textbf{E}_3$. This procedure allows to refine the integrated query features with added attention around the area of both hands and object in the next step. 
    \item Query Feature Update: In the ``Query Feature Update Block", integrated query features are further refined using cross-attention (with updated $\textbf{E}_3$ as keys and values) and self-attention layers. This block allows to capture the fine-grained details such as finger joints, palm surface characteristics, object contact points, and specific regions on the hand and object that contribute to a detailed understanding of their poses. Leveraging the refined query features, we proceed to estimate target classes, hand poses, and object poses.
\end{itemize}

\noindent \textbf{Prediction}. On top of the refined query features at each decoder layer, we apply three 3-layer MLPs and a linear layer to output fine-grained 2D keypoints for left and right hands, 3D object poses and target classes, respectively. \aaai{Notably, the same sets of linear layers with shared weights were employed to estimate coase 2D keypoints for hands \& object and coarse classification in the previous step in the decoder. 2D keypoints of object were extracted from estimated 3D keypoints before further processing in the Location-based Feature Extraction module. } 

\noindent \new{Motivated by the performance and high efficiency of graph-oriented attention in 3D hand pose estimation \cite{zhao2022graformer},} by leveraging the skeleton structure of hands to be as graph-structured data, we further refine hand poses by passing the 2D input keypoints through a series of keypoint attention and ChebGConv \cite{zhao2022graformer} layers. The keypoint attention block integrates multi-head attention and graph convolution layers, while the ChebGConv block incorporates Chebyshev graph convolutional layers. This strategic combination exploits the inherent connectivity among keypoints, enabling the model to effectively capture and estimate 3D poses from 2D coordinates. This methodology proves to be a robust and \new{highly efficient} solution in~\cite{zhao2022graformer} for overcoming the inherent challenges associated with direct 3D pose prediction from 2D feature maps. 
\subsection{Loss Function} \label{subsec:loss-function}
\begin{table}[t!]
    \begin{minipage}[t!]{\linewidth}
    \centering
    \resizebox{\linewidth}{!}{
    \begin{tabular}{l|ccc|cc}
    \hline
    \multirow{2}{*}{Method}                                             & \multicolumn{3}{c|}{H2O}                                                                   & \multicolumn{2}{c}{FPHA}                          \\ \cline{2-6} 
                                                                        & \multicolumn{1}{c|}{Left}         & \multicolumn{1}{c|}{Right}        & Obj. (L/R)         & \multicolumn{1}{c|}{Right}       & Obj.        \\ \hline
    Hasson~\etal.~\shortcite{hasson2020leveraging}    & \multicolumn{1}{c|}{39.6}          & \multicolumn{1}{c|}{41.9}          & 67.5/66.1    & \multicolumn{1}{c|}{18.0}          & 22.3          \\ 
    Tekin~\etal.~\shortcite{tekin2019h+}                     & \multicolumn{1}{c|}{41.4}          & \multicolumn{1}{c|}{38.9}          & 48.1/52.6    & \multicolumn{1}{c|}{15.8}          & 24.9          \\ 
    Kwon~\etal.~\shortcite{kwon2021h2o}                & \multicolumn{1}{c|}{41.5}          & \multicolumn{1}{c|}{37.2}          & 47.9          & \multicolumn{1}{c|}{-}             & -             \\ 
    Wen~\etal.~\shortcite{wen2022hierarchical}         & \multicolumn{1}{c|}{35.0}          & \multicolumn{1}{c|}{36.1}          & -              & \multicolumn{1}{c|}{15.8}          & -             \\ 
    Aboukhadra1~\etal.~\shortcite{aboukhadra2022thor} & \multicolumn{1}{c|}{36.8}          & \multicolumn{1}{c|}{36.5}          & 73.9          & \multicolumn{1}{c|}{-}             & -             \\ 
    Cho~\etal.~\shortcite{cho2023transformer}                                                                & \multicolumn{1}{c|}{24.4} & \multicolumn{1}{c|}{25.8} & 45.2 & \multicolumn{1}{c|}{15.0} & 21.0 \\ \hline
    Ours                                                                & \multicolumn{1}{c|}{\textbf{20.1}} & \multicolumn{1}{c|}{\textbf{19.9}} & \textbf{32.9} & \multicolumn{1}{c|}{\textbf{14.2}} & \textbf{18.8} \\ \hline
    \end{tabular}
        }
        \caption{Mean End-Point Error comparison across SOTA pose estimation pipelines have been shown. Experiments are performed on test sets of H2O and FPHA datasets. Single-hand methodologies~\cite{hasson2019learning, tekin2019h+} are tested for left and right hand object interactions separately. Our method outperforms others by a significant margin. Best results are in bold.}
        \label{table:pose_results}
    \end{minipage}
    % \begin{minipage}[t]{.25\linewidth}
    % \centering
    %     \resizebox{\linewidth}{!}{
    %     \begin{tabular}{c|c}
    %     \hline
    %     \multirow{2}{*}{Method}                                       & H2O                   \\ \cline{2-2}
        
    %                                                                     & Acc        \\ 
    %     \hline
    %     C2D~\shortcite{wang2018non}                    & 70.7                        \\ 
    %     I3D~\shortcite{carreira2017quo}                & 75.2                        \\ 
    %     SlowFast~\shortcite{feichtenhofer2019slowfast} & 77.7                        \\ 
    %     Tekin~\etal.~\shortcite{tekin2019h+}          & 68.9                    \\ 
    %     Kwon~\etal.~\shortcite{kwon2021h2o}           & 79.3                        \\ 
    %     Wen~\etal.~\shortcite{wen2022hierarchical}     & 86.4                        \\  
    %     Cho~\etal.~\shortcite{cho2023transformer}                                                            & 90.9  \\ \hline
    %      Ours                                  & \textbf{91.3} \\ \hline
    %     \end{tabular}
    %     }
    %    \caption{
    %     \new{Comparison of Top-1 Accuracy for Hand-Object Interaction Recognition in H2O dataset.} 
    %     }
    %     \label{table:action_results}
    % \end{minipage}
\end{table}
\begin{table}[t!]
    \centering
    \begin{minipage}[t]{\linewidth}
    \centering
        \resizebox{\linewidth}{!}{
        \begin{tabularx}{\textwidth}{X|c}
        \hline
        \multirow{2}{*}{Method}                                       & H2O                   \\ \cline{2-2}
        
                                                                        & Accuracy        \\ 
        \hline
        C2D~\cite{wang2018non}                    & 70.7                        \\ 
        I3D~\cite{carreira2017quo}                & 75.2                        \\ 
        SlowFast~\cite{feichtenhofer2019slowfast} & 77.7                        \\ 
        Tekin~\etal.~\shortcite{tekin2019h+}          & 68.9                    \\ 
        Kwon~\etal.~\shortcite{kwon2021h2o}           & 79.3                        \\ 
        HTT~\cite{wen2022hierarchical}     & 86.4                        \\  
        H2OTR~\cite{cho2023transformer}                                                            & 90.9  \\ \hline
         Ours                                  & \textbf{91.3} \\ \hline
        \end{tabularx}
        }
       \caption{
        \new{Comparison of Top-1 Accuracy for Hand-Object Interaction Recognition in H2O dataset.} 
        }
        \label{table:action_results}
    \end{minipage}
\end{table}

The overall loss function $\mathcal{L}$ for training \tool can be written as follows:
\begin{eqnarray}
\mathcal{L} = \lambda_{\text{CE}}\mathcal{L}_{\text{CE}} + \lambda_{\text{KP}}  \mathcal{L}_{\text{KP}} + \lambda_{\text{QP}}  \mathcal{L}_{\text{QP}} + \lambda_{\text{CM}}  \mathcal{L}_{\text{CM}}  
\end{eqnarray}
where, $\mathcal{L}_{\text{CE}}$ is the classification loss for the final classification head, $\mathcal{L}_{\text{QP}}$ is the query proposal loss,~$\mathcal{L}_{\text{CM}}$ is the contact map estimation loss and $\mathcal{L}_{\text{KP}}$ is the loss for estimating keypoints of two hands and an object. $\lambda_{\text{CE}}$, $\lambda_{\text{QP}}$, $\lambda_{\text{CM}}$ and $\lambda_{\text{KP}}$ are the hyper-parameters to balance the weights of losses during the training.

\noindent \textbf{Classification Loss.} For classification loss, $\mathcal{L}_{\text{CE}}$,  of our model, similar to \cite{cho2023transformer} we use Hungarian algorithm \cite{kuhn1955hungarian} to match our predicted output classes, $\bm{C}'$ to the ground truth classes, $\bm{C}$. The output of this classification head forms a probability simplex for each class, encompassing left hand, right hand, and individual object categories. To train the model effectively, we apply cross-entropy loss, aiming to maximize the likelihood of the true class occurrences within the scene.

\noindent \textbf{Keypoints Estimation Loss.} We apply L1 loss between predicted keypoints and ground-truth keypoints as follows:
\begin{eqnarray}
\mathcal{L}_{\text{KP}} = \lVert \bm{K}_{\text{h}}^{\text{3D}} - \bm{K}_{\text{h}^\text{*}}^{\text{3D}} \rVert_1 &+ \lVert \bm{K}_{\text{h}}^{\text{2D}} - \bm{K}_{\text{h}^\text{*}}^{\text{2D}} \rVert_1  &\\ \nonumber&+ \lVert \bm{K}_{\text{o}}^{\text{3D}} - \bm{K}_{\text{o}^\text{*}}^{\text{3D}} \rVert_1&
\end{eqnarray}
where $\bm{K}_{\text{h}}^{\text{3D}} \in \mathbb{R}^{2 \times 21 \times 3}$, $\bm{K}_{\text{h}}^{\text{2D}} \in \mathbb{R}^{2 \times 21 \times 2}$ and $\bm{K}_{\text{o}}^{\text{3D}} \in \mathbb{R}^{21 \times 3}$ denote ground truth 3D hand keypoints, 2D hand keypoints and 3D object keypoints, respectively. And, $\bm{K}_{\text{h}^\text{*}}^{\text{3D}}$, $\bm{K}_{\text{h}^\text{*}}^{\text{2D}}$ and $\bm{K}_{\text{o}^\text{*}}^{\text{3D}}$  represent the estimated 3D hand keypoints, 2D hand keypoints and 3D object keypoints, respectively.

\noindent \textbf{Query Proposal Loss.}
We use the binary cross-entropy loss for left and right hand \new{query proposal modules} and cross-entropy loss for supervising object \new{query proposal module}. We apply binary classification loss to the left and right hand \new{query proposal modules} as for both of them, the objective is to classify hand or ``no object''. For each query, we find the optimal match using the Hungarian algorithm \cite{kuhn1955hungarian} by computing classification and location costs (\ie, whether the estimated query position is in the region of the object of interest or not). To make sure queries from each segment are focused on a different area of the region of interest, we do not match with another query location to a previously matched ground truth location. 
The query proposal loss is defined as follows,
\begin{eqnarray} \mathcal{L}_{\text{QP}} = \mathcal{L}_{\text{qo}} +  \mathcal{L}_{\text{ql}} + \mathcal{L}_{\text{qr}}
\end{eqnarray}
where $\mathcal{L}_{\text{ql}}$, $\mathcal{L}_{\text{qr}}$, $\mathcal{L}_{\text{qo}}$ are proposal losses for the left and right hand and an object \new{query proposal modules}, respectively. Our method often fails to distinguish between the highly similar features of the left and right hands. To address the challenge and increase the likelihood of capturing features for both hands, we segment a set of queries for each specific region—left hand, right hand, and object and to channel each set of queries towards its corresponding area of interest, we employ the respective query proposal losses. 

\noindent \textbf{Contact Map Estimation Loss.} To guide our model to estimate the ground truth contact map, we obtain the left-hand contact map, $\textbf{C}_\text{m}^\text{left} \in \mathbb{R}^{778 \times 1}$ and right-hand contact map, $\textbf{C}_\text{m}^{\text{right}} \in \mathbb{R}^{778 \times 1}$ from hands and object meshes by following \cite{cho2023transformer}. For our task, we combine the contact map of both hands to obtain the ground truth contact map, $\textbf{C}_\text{M} \in \mathbb{R}^{2 \times 778 \times 1}$. For the loss, we calculate the $\text{L1}$ loss between the predicted contact map, $\textbf{C}_\text{M}^{\prime} \in \mathbb{R}^{2 \times 778 \times 1}$, and the ground truth contact map, $\textbf{C}_\text{M}$. Therefore, the contact map estimation loss, $\mathcal{L}_{\text{CM}}$ can be defined as follows,
\begin{eqnarray} \mathcal{L}_{\text{CM}} = \lVert \textbf{C}_\text{M} - \textbf{C}_{\text{M}}^{\prime}\rVert_1.
\end{eqnarray}
\begin{figure}[t!]
  \begin{subfigure}[t]{0.32\linewidth}
    \centering
    \includegraphics[width=\linewidth]{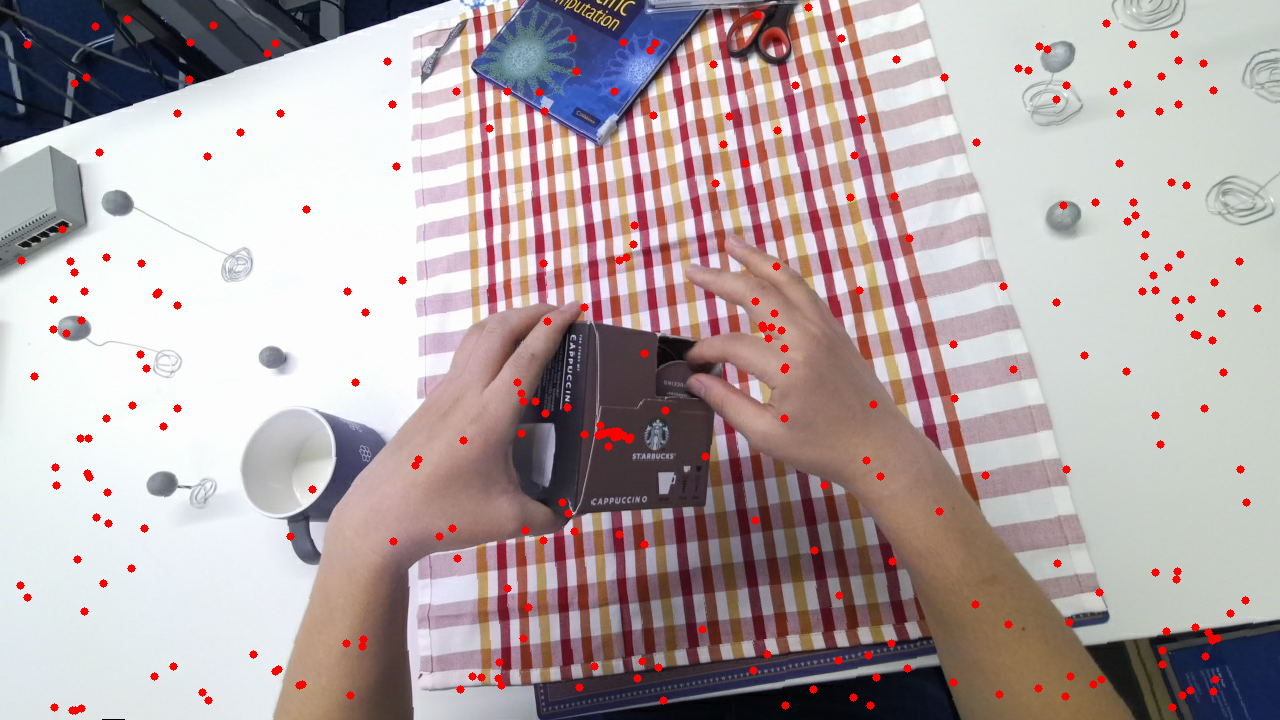}
    \label{fig:h2otr}
  \end{subfigure}
  \begin{subfigure}[t]{0.32\linewidth}
    \centering
    \includegraphics[width=\linewidth]{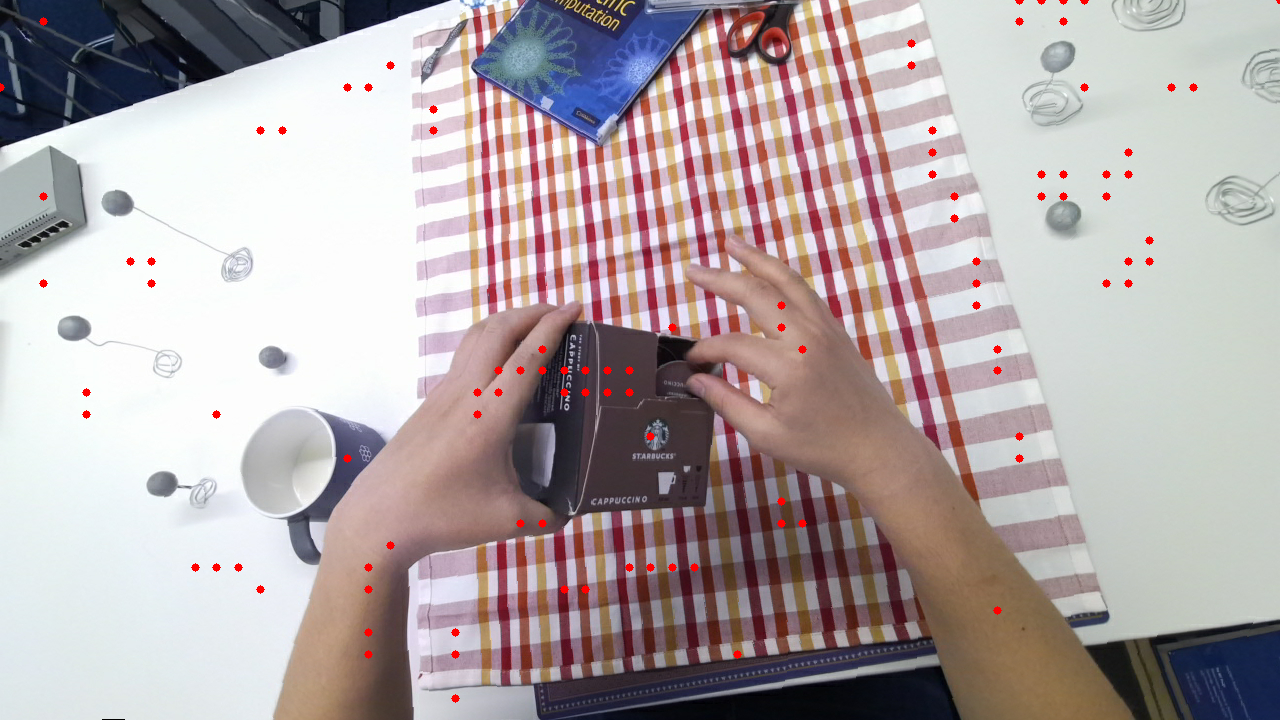}
    \label{fig:non_sg_queries}
  \end{subfigure} 
  \begin{subfigure}[t]{0.32\linewidth}
    \centering
    \includegraphics[width=\linewidth]{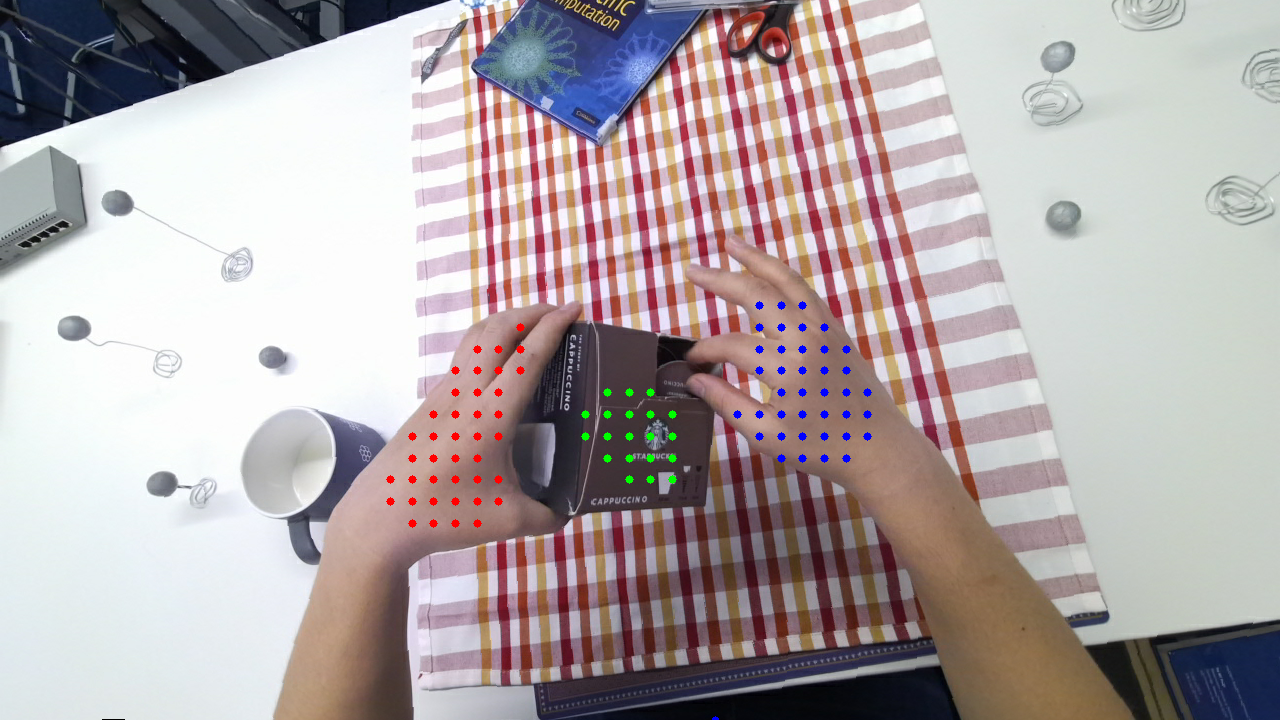}
    \label{fig:ours}
  \end{subfigure}
   \caption{Query location visualization: (a) \textbf{Left:} query locations of H2OTR \cite{cho2023transformer}, employing 300 queries. Notably, a substantial amount of queries are distributed in backgrounds. (b) \textbf{Middle:} Our hand-object query locations w/o Query division block. Due to feature similarities between two hands, a considerable number of queries concentrate on the left hand than the right hand, which reduces the accuracy of the right hand. (c) \textbf{Right:} Our hand-object query locations. Queries for left and right hands are highlighted in red and blue, respectively. Queries for objects are denoted as green. The query proposal loss ensures that each query concentrates on its specific region of interest.}
  \label{fig:queries}
\end{figure}
\section{Experiments}
\subsection{Datasets and Evaluation Metrics} \label{subsec:datasetsneval}
We conducted evaluations on two distinct datasets: H2O~\cite{kwon2021h2o} and FPHA~\cite{garcia2018first}, both of which include annotations for 3D hand poses, object 6D poses, object types and interaction classes. For hand-object pose estimation, we measure the mean end-point error (in mm) across 21 joints, and for our extended experiment interaction recognition, we use top-1 accuracy with an off-the-shelf network \cite{cho2023transformer}. Further details on datasets, implementation details and evaluation metrics are available in our supplemental materials.
% \begin{table}[t!]
% \begin{minipage}[t]{\linewidth}
% \centering
% \resizebox{\linewidth}{!}{
% \begin{tabular}{l|l|c|c|c}
% \hline
% \multicolumn{2}{c|}{-}            & w/o  HOQ & w/ HOQ & w/ HOQ + CM \\ \hline
% \multirow{3}{*}{H2O}  & Left  & 27.3                   & 22.6                 & \textbf{20.1}                                        \\ 
%                       & Right & 33.5                   & 21.4                 & \textbf{19.9}                                        \\  
%                       & Object  & 36.5                   & 33.1                 & \textbf{32.9}                                        \\ \hline
% \multirow{2}{*}{FPHA} & Right & 22.3                   & 15.8                & \textbf{14.2}                                        \\ 
%                       & Object  & 24.2                   & 20.3                 & \textbf{18.8}                                        \\ \hline
% \end{tabular}
% }
% \caption{Ablation studies on different components of our framework. In the table, HOQ denotes hand-object queries and CM denotes contact map features.}
% \label{table:ablation}
% \end{minipage}\hfill
% \end{table}

\begin{table}[t!]
   \begin{minipage}[t]{\linewidth}
\centering
\resizebox{\linewidth}{!}{
\begin{tabular}{l|c|c|c|c}
\hline
\multirow{2}{*}{Model}      & \multicolumn{3}{c|}{H2O}   & \multirow{2}{*}{FPS}                                                                \\ \cline{2-4}  
           & \multicolumn{1}{l|}{Left} & \multicolumn{1}{l|}{Right} & \multicolumn{1}{l|}{Object} \\ \hline
          % Query feat. update 
          Ours w/o EFU \& LFE & 26.4                        & 25.8                          & 36.1    &     \textbf{58.2}          \\
Ours w/o LFE & 22.3                         & 21.8                          & 33.9       &      56.5 \\ \hline
Ours & \textbf{20.1}                         & \textbf{19.9}                          & \textbf{32.9}       &      53.5 \\ \hline
\end{tabular}
}
\caption{Ablation studies on each component of QORT Transformer decoder. EFU denotes Enhanced Feature update and LFE denotes location-based feature extraction.
}
\label{table:pixel_update}
\end{minipage}\hfill
\end{table}
\subsection{Experiment Results} 
\begin{figure*}[t!]
\centering
\includegraphics[width=\linewidth]{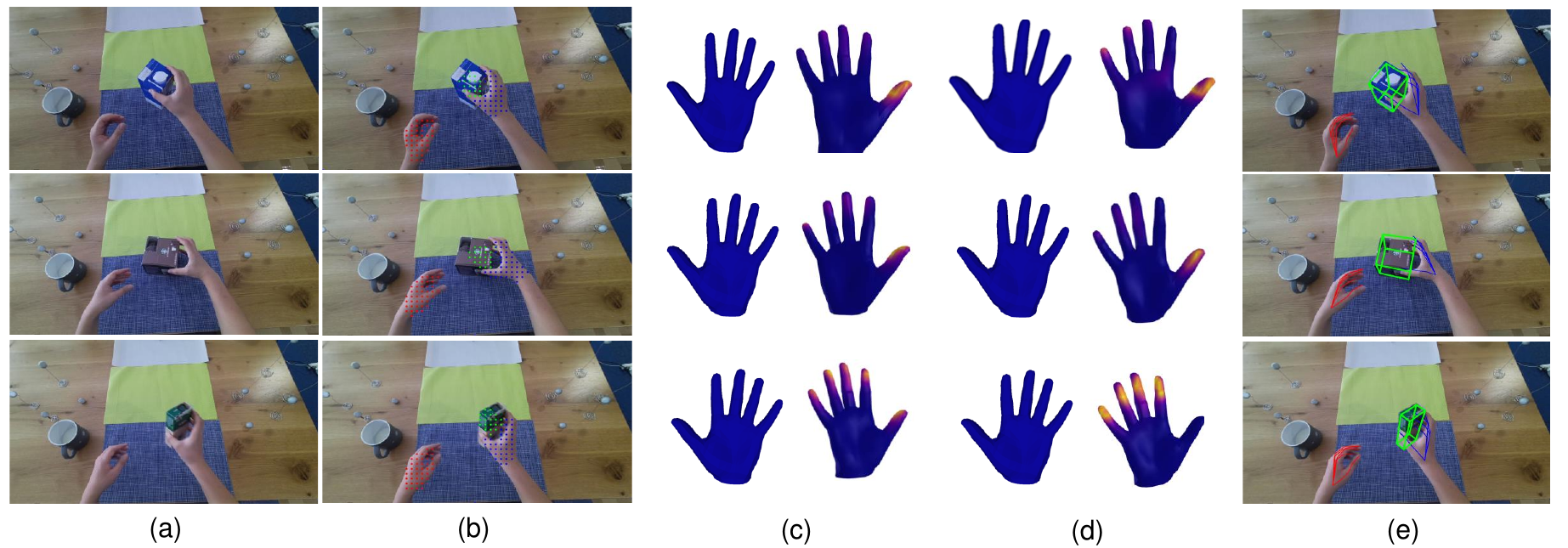}
\caption{Examples of estimated 3D poses on H2O dataset: For a separate example in each row, the figure represents (a) input RGB image, (b) our hand-object queries, (c) ground-truth contact map, (d) predicted contact map, and (e) final 3D pose estimation results, respectively.}
\label{fig:h2o_results}
\end{figure*}

\begin{figure*}[t!]
% \captionsetup{font=footnotesize}
\centering
\includegraphics[width=\linewidth]{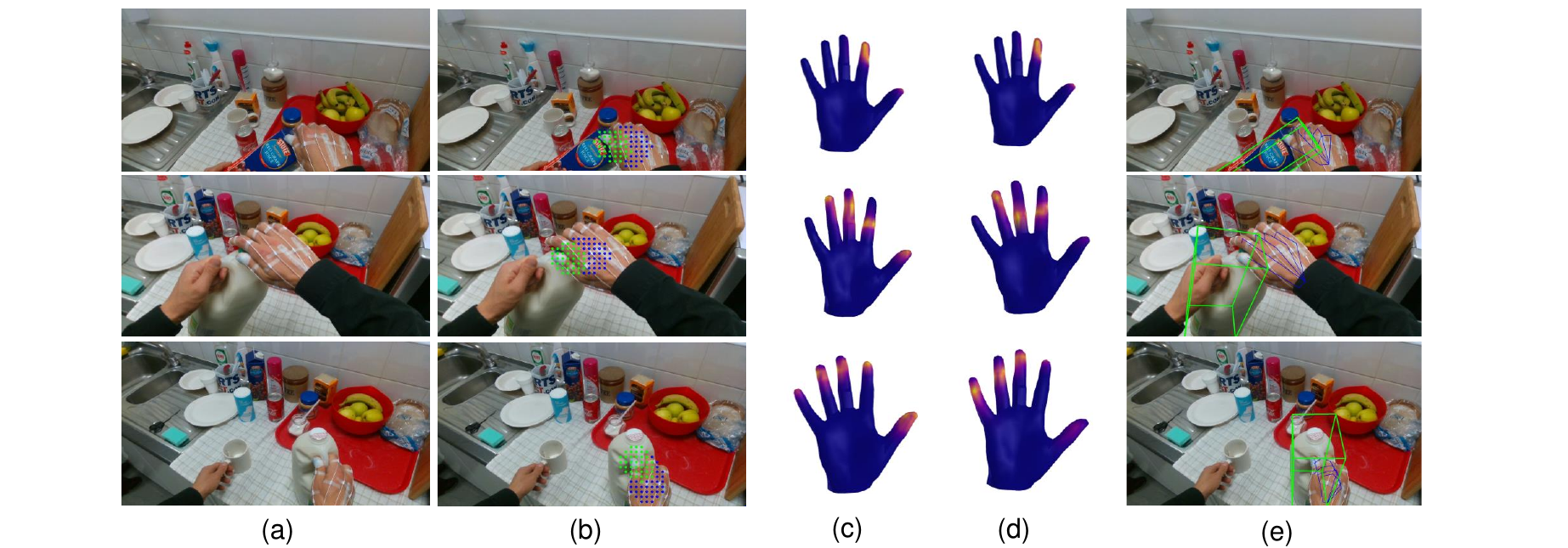}
\caption{Examples of estimated 3D poses on FPHA dataset. For a separate example in each row, the figure represents (a) input RGB image, (b) our hand-object queries, (c) ground-truth contact map, (d) predicted contact map, and (e) final 3D pose estimation results, respectively.}
\label{fig:fpha_results}
\end{figure*}

\textbf{Pose Estimation.} We compare our method with SOTA hand-object pose estimation methods that use a single RGB image as input on the H2O and FPHA datasets. As experiment results in Table~\ref{table:pose_results} demonstrate, our method achieves SOTA results and outperforms all previous methods by a significant margin. \new{Compared to \citeauthor{cho2023transformer}, we achieve substantial gains on the H2O dataset: 17.6\% for the left hand, 22.8\% for the right hand, and 27.2\% for the object. On the FPHA dataset, our method outperforms the state-of-the-art by a decisive margin: 5.3\% for the right hand and 10.4\% for the object.} 
\citeauthor{hasson2020leveraging} and Tekin~\etal~\cite{tekin2019h+} predict pose for only a single hand. \citeauthor{wen2022hierarchical}'s method does not predict object poses. Figs.~\ref{fig:h2o_results} and~\ref{fig:fpha_results} show the example estimated 3D poses of hands and an object on H2O and FPHA datasets, respectively.\\
\aaai{\textbf{Interaction Recognition.} In our extended experiment on hand-object interaction recognition, we utilized the action recognition module from \citeauthor{cho2023transformer} to assess the impact of our improved inference speed of our pose estimator on interaction classification. By replacing their pose estimator with our proposed method, \tool, we observed an increase in performance from 90.9 to 91.3 on the H2O dataset, as shown in Table \ref{table:action_results}. Additionally, incorporating our method led to a significant boost in inference speed, achieving 53.3 FPS compared to the 24.97 FPS of Cho et al.'s framework on an RTX 3090TI GPU.}

\begin{table}[t!]
\begin{minipage}[t]{\linewidth}
\centering
\resizebox{\linewidth}{!}{
\begin{tabular}{l|l|c|c|c}
\hline
\multicolumn{2}{c|}{-}            & w/o  HOQ & w/ HOQ & w/ HOQ + CM \\ \hline
\multirow{3}{*}{H2O}  & Left  & 27.3                   & 22.6                 & \textbf{20.1}                                        \\ 
                      & Right & 33.5                   & 21.4                 & \textbf{19.9}                                        \\  
                      & Object  & 36.5                   & 33.1                 & \textbf{32.9}                                        \\ \hline
\multirow{2}{*}{FPHA} & Right & 22.3                   & 15.8                & \textbf{14.2}                                        \\ 
                      & Object  & 24.2                   & 20.3                 & \textbf{18.8}                                        \\ \hline
\end{tabular}
}
\caption{Ablation studies on different components of our framework. In the table, HOQ denotes hand-object queries and CM denotes contact map features.}
\label{table:ablation}
\end{minipage}\hfill
\end{table}
\subsection{Ablation Study} \label{subsec:ablation}
In this section, we conduct ablation studies on our improved queries and decoder of our proposed model, \tool. Further ablations of other configurations of our architecture are available in our supplementary mat. Each component is evaluated on the test sets of the H2O and/or FPHA datasets. \\
\noindent \aaai{\textbf{Analysis of Our Improved Queries.} To enhance the efficiency of our QORT Transformer decoder, we introduce hand-object queries combined with a query proposal loss, aimed at improving query location accuracy by ensuring queries focus on regions of interest, such as the left and right hands and the interacting object. This approach addresses the challenge of imprecise query location estimation, which often arises in hand pose estimation due to the semantic similarity between the left and right hands. The improved query distribution leads to a more balanced model performance across both hands, as evidenced by a substantial performance increase demonstrated in Table \ref{table:ablation}
 and illustrated in Figure \ref{fig:queries}. Additionally, to further refine the model's accuracy in interacting hands and object pose estimation, we incorporate contact map features that capture the spatial relationships between the hands and the object. By adding these features to the query inputs, our model achieves state-of-the-art results in hand and object pose estimation from a single RGB image, as shown in Table \ref{table:ablation}.} \\
\noindent \aaai{\textbf{Analysis on QORT Transformer Decoder.}  We adopted a three-step update approach in our proposed decoder to compensate for the reduced feature maps compared to heavier architectures \cite{cho2023transformer, Cheng_2022_CVPR}. Unlike \cite{Cheng_2022_CVPR}, which relies solely on query feature updates, we first refine enhanced features $\textbf{E}_3$ using cross-attention and then apply location-based feature extraction to focus on hands and objects. This refined $\textbf{E}_3$ then, guides the final query feature update. Although this method slightly reduces FPS, it improves overall pose estimation performance, as shown in Table \ref{table:pixel_update}.}
\section{Conclusion} \label{sec:conclusion} 
% In this work, we present \tool, the first real-time Transformer-based framework for two-hand and object pose estimation. Our lightweight feature decoder with pyramid pooling reduces queries to 108 and integrates contact map features for hand-object interactions. Using a three-step update strategy, \tool minimizes encoder use and limits the architecture to one decoder. It achieves state-of-the-art performance on H2O and FPHA datasets, running at 53.5 FPS on an RTX 3090TI, significantly advancing real-time pose estimation.
In this work, we present \tool, the first real-time Transformer-based framework designed specifically for two hands and interacting object pose estimation. Our approach introduces a lightweight feature decoder with pyramid pooling, significantly reducing the number of queries to just 108 while effectively incorporating contact map features to model intricate hand-object interactions. By leveraging a novel three-step update strategy (Enhanced Feature Update, Location-based Enhancement, Query Feature Update, \tool minimizes the computational overhead of the encoder and simplifies the architecture by utilizing a single decoder. These innovations collectively enable \tool to achieve state-of-the-art performance on widely-used hand-object interaction benchmarks such as the H2O and FPHA datasets. Furthermore, \tool operates at an impressive speed of 53.5 frames per second (FPS) on an RTX 3090TI GPU, demonstrating its practical viability for real-time applications. The combination of accuracy, efficiency, and real-time performance positions \tool as a significant advancement in the field of hands-object pose estimation, paving the way for new possibilities in applications, such as human-computer interaction, robotics, augmented reality and virtual reality.
\section{Acknowledgements}
 This work was supported by IITP grants (No. RS-2020-II201336 Artificial intelligence graduate school program (UNIST) 10\%; No. RS-2021-II212068 AI innovation hub 10\%; No. RS-2022-II220264 Comprehensive video understanding and generation with knowledge-based deep logic neural network 20\%) and the NRF grant (No. RS-2023-00252630 20\%), all funded by the Korean government (MSIT). This work was also supported by Korea Institute of Marine Science \& Technology Promotion (KIMST) funded by Ministry of Oceans and Fisheries (RS-2022-KS221674) 20\% and received support from AI Center, CJ Corporation (20\%).
\bibliography{aaai25}
\end{document}